%% file: main.tex
\documentclass{article}
\usepackage{spconf,amsmath,graphicx}
\usepackage{graphicx}
\usepackage{caption} 
\usepackage{subcaption}
\usepackage{xspace}
\usepackage{enumitem}
\usepackage{multirow}
\usepackage{hyperref}

\title{Hydra-Bench: A Benchmark for Multi-Modal Leaf Wetness Sensing}
\name{Yimeng Liu, Maolin Gan, Yidong Ren, Gen Li, Jingkai Lin, Younsuk Dong,  Zhichao Cao}
\address{Michigan State University, East Lansing, United States of America}
\begin{document}
%
\maketitle
\begin{abstract}
Leaf wetness detection is a crucial task in agricultural monitoring, as it directly impacts the prediction and protection of plant diseases.
However, existing sensing systems suffer from limitations in robustness, accuracy, and environmental resilience when applied to natural leaves under dynamic real-world conditions. 
To address these challenges, we introduce a new multi-modal dataset specifically designed for evaluating and advancing machine learning algorithms in leaf wetness detection. 
Our dataset comprises synchronized mmWave raw data, Synthetic Aperture Radar (SAR) images, and RGB images collected over six months from five diverse plant species in both controlled and outdoor field environments.
We provide detailed benchmarks using the Hydra \cite{hydra24liu} model, including comparisons against single modality baselines and multiple fusion strategies, as well as performance under varying scan distances.
Additionally, our dataset can serve as a benchmark for future SAR imaging algorithm optimization, enabling a systematic evaluation of detection accuracy under diverse conditions.

\end{abstract}
\input{content/1_introduction}

\input{content/2_problem}
\input{content/3_dataset}
\input{content/4_testing}

\input{content/5_evaluation}

\input{content/6_conclusion}







\bibliographystyle{IEEEbib}
\bibliography{refs}

\end{document}

%% file: content/1_introduction.tex
\section{Introduction}
\label{sec:intro}

Agriculture is a significant part of the global economy, accounting for approximately 4\% of global GDP and exceeding 25\% in some developing nations~\cite{worldbankAgri}.
However, the increasing frequency and severity of plant diseases show significant threats to agricultural productivity, food security, and biodiversity ~\cite{Singh2023Climate}. 
A major factor underlying disease development is leaf wetness, which is the presence of water on leaf surfaces.
It can be the result of dew, precipitation, fog, or irrigation \cite{plants12152800}.
The duration of leaf wetness is particularly critical, as it can promote the growth of various pathogens, including Venturia inaequalis, etc~\cite{huber1992modeling}.
As a result, accurate detection of leaf wetness duration (LWD) is essential for effective monitoring and control of plant diseases~\cite{rowlandson2015reconsidering}.
Previous studies have demonstrated the importance of accurately detecting LWD to protect crop yields in various species, including strawberries~\cite{mackenzie2012use}, grapes~\cite{gubler1999control}, and lettuce~\cite{wu2001validation}.

To improve the performance of LWD detection, researchers have explored various sensing modalities~\cite{duvdevani1947optical, gan2023poster, PHYTOS31, nguyen2023bio}.
However, existing systems still face significant limitations related to sensing accuracy, environmental robustness, and system efficiency.
The leaf wetness sensors (LWS) utilize synthetic leaves~\cite{PHYTOS31,  nguyen2023bio}, which differ from real leaves in size, shape, and material properties. 
These errors can result in detection errors of up to 30 minutes.
RGB imaging approaches~\cite{duvdevani1947optical} are highly susceptible to variations in lighting.
In addition, mmWave-based techniques~\cite{gan2023poster} are sensitive to leaf movement caused by wind. 
Also, it required time-consuming scanning procedures, which further compromised system efficiency.

\begin{figure}[btp]

\centering
\centerline{\includegraphics[width=8cm]{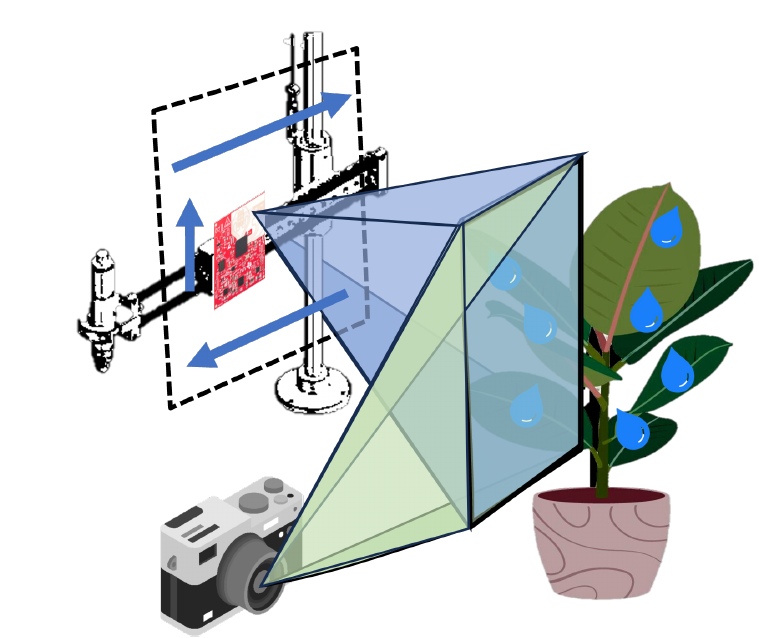}}

 \caption{Leaf Wetness Detection with RGB camera and SAR.}
\label{fig:teaser}
\end{figure}

In this paper, we present the dataset used for training and testing Hydra \cite{hydra24liu}, shown in Figure \ref{fig:teaser}.
Hydra is the first contactless multi-modality sensing system explicitly designed for accurate LWD detection. 
Built using commercial off-the-shelf hardware, it integrates mmWave and RGB imaging to enable direct, high-fidelity scanning of plant surfaces. 
Additionally, Hydra combines deep learning with advanced multi-modal data fusion, effectively overcoming challenges associated with aligning data across different spatial and temporal resolutions. 
It achieves a high detection accuracy of 96\% in controlled indoor scenarios and approximately 90\% in challenging outdoor farm environments, including rainy, dawn, and low-light night time. 
In particular, Hydra reduces the LWD detection error margin to just 2 minutes, significantly outperforming prior approaches that relied on synthetic leaves or single-modality sensing.

Our dataset comprises five distinct plant species, collected over more than six months to capture a wide range of growth patterns, spatial arrangements, and foliage distributions, thus enhancing the diversity of the dataset. 
The data include both indoor and outdoor environmental conditions to ensure robustness across deployment scenarios. 
Each sample consists of raw mmWave data, Synthetic Aperture Radar (SAR) imaging, and an RGB image, all carefully calibrated according to the protocol established in~\cite{hydra24liu}.
The dataset can be found at \url{https://drive.google.com/drive/folders/1C0mq5vZgEJOYMNL1vSghvFn0OfyIi8Cm}.
This dataset can serve as a benchmark to inspire future research on multimodal fusion and optimization of the SAR imaging algorithm.

%% file: content/2_problem.tex
\section{Understanding the Problem}
\label{sec: problem}

\subsection{mmWave Sensing}
mmWave utilizes electromagnetic waves with wavelengths ranging from 1 to 10 millimeters. 
This short wavelength endows the mmWave radar with high sensitivity to fine surface textures, making it well-suited for detecting subtle changes such as leaf wetness.
One of mmWave's key advantages lies in its responsiveness to material properties.
Materials reflect mmWave signals differently based on their permittivity, a physical property that influences how electromagnetic waves propagate through them. 
Water, which has significantly higher permittivity than dry leaf tissue, alters the reflection characteristics of wet leaf surfaces. 
This contrast in reflective behavior enables mmWave systems to differentiate between wet and dry leaves effectively \cite{hydra24liu, gan2023poster, Adonis25Liu, Proteus25Liu}, providing a powerful, contactless modality for leaf wetness detection.

\subsection{Synthetic Aperture Radar Imaging}
\begin{figure}[t!]

\begin{minipage}[b]{.48\linewidth}
  \centering
  \centerline{\includegraphics[width=4.0cm]{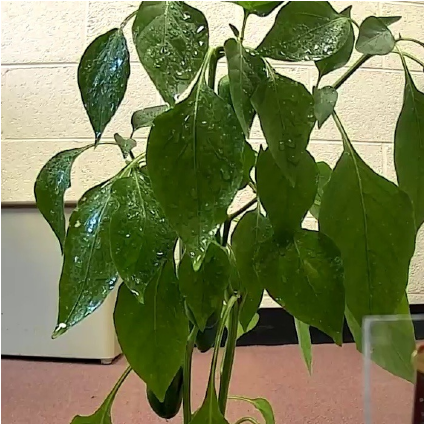}}
  \centerline{RGB Image}\medskip
\end{minipage}
\begin{minipage}[b]{.48\linewidth}
  \centering
  \centerline{\includegraphics[width=4.0cm]{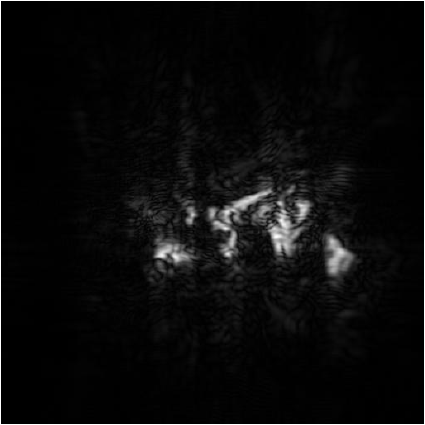}}
  \centerline{SAR Image}\medskip
\end{minipage}
\caption{Example of the dataset with RGB and SAR image}
\label{fig:sar-example}
\end{figure}

SAR is a well-established technique in radar systems.
It can generate high-resolution images by synthesizing a large aperture through the relative motion between the radar and the target. 
By effectively expanding the aperture size, SAR enhances the radar's ability to capture fine-grained spatial details, which is critical for accurately identifying subtle features such as leaf wetness.
The example is shown in Figure \ref{fig:sar-example}.
When combined with frequency-modulated continuous-wave (FMCW) chirps, the mmWave radar becomes highly sensitive to range variations between the radar and the target. 
This capability enables SAR to image multiple depth layers, providing cross-sectional views that offer a rich spatial context.

\subsection{Leaf Wetness Detection}

Leaf wetness is defined as the presence of water on the leaf surface.
It is a simple but critically important factor for agricultural health monitoring. 
The primary objective in detecting LWD is to accurately classify whether a leaf surface is wet or dry over time.
From a machine learning perspective, this problem introduces key questions around trustworthy classification: What specific features distinguish a wet leaf from a dry one?
Is it the spectral signature of water, micro-textural cues on the surface, or some other latent factors? 
These ambiguities complicate not only model training but also the interpretability of the decision process, posing a barrier for deploying models with confidence in agricultural applications.

Our dataset addresses these challenges by including diverse data collected under varying environmental conditions. 
Each sample pairs high-resolution RGB images with mmWave SAR data, both precisely calibrated, providing a rich multi-modal foundation for robust and explainable model development. 
This dataset enables researchers to benchmark SAR imaging algorithms and multi-modality fusion strategies. 
It supports evaluation across key dimensions, including accuracy, resilience, generalization, and interoperability. 
For both machine learning practitioners and agricultural technologists, this dataset offers a high-quality resource that bridges the gap between controlled experiments and real-world deployment. 
It is particularly well-suited for training deep learning models, optimizing SAR-based imaging pipelines, and advancing multi-modal fusion methods. 
Furthermore, it opens new opportunities for research in explainable AI, establishing a valuable benchmark for future innovations in environmental sensing and precision agriculture.

%% file: content/3_dataset.tex
\section{Dataset}
\label{sec: dataset}

\subsection{Implementation}
\begin{figure}[!t]
   \centering
   \begin{subfigure}[b]{0.23\textwidth}
       \centering\includegraphics[width=\textwidth]{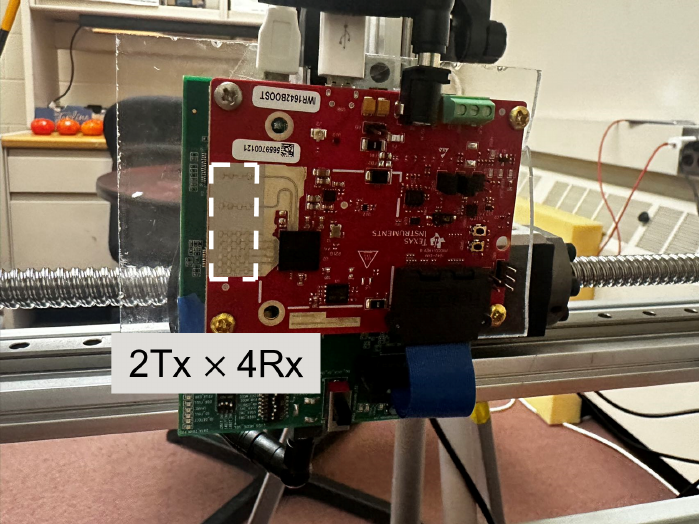}
   \caption{mmWave Radar}\label{fig:radar} 
   \end{subfigure}
   \begin{subfigure}[b]{0.23\textwidth}
       \centering\includegraphics[width=\textwidth]{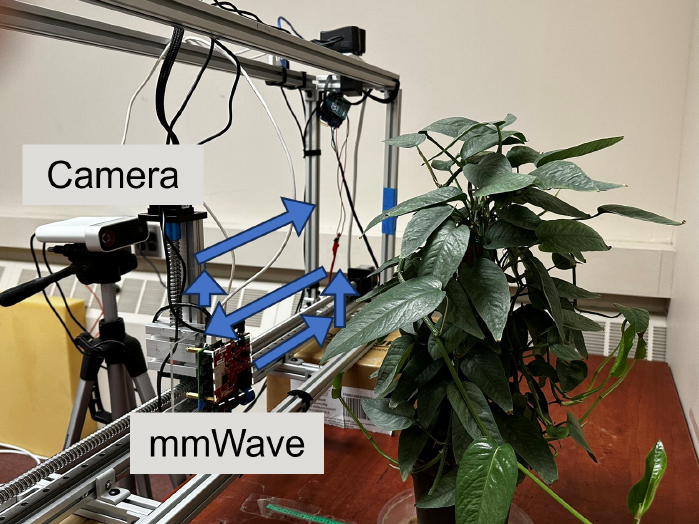}
   \caption{Indoor Setup}\label{fig:indoor}
   \end{subfigure}
   \begin{subfigure}[b]{0.23\textwidth}
       \centering\includegraphics[width=\textwidth]{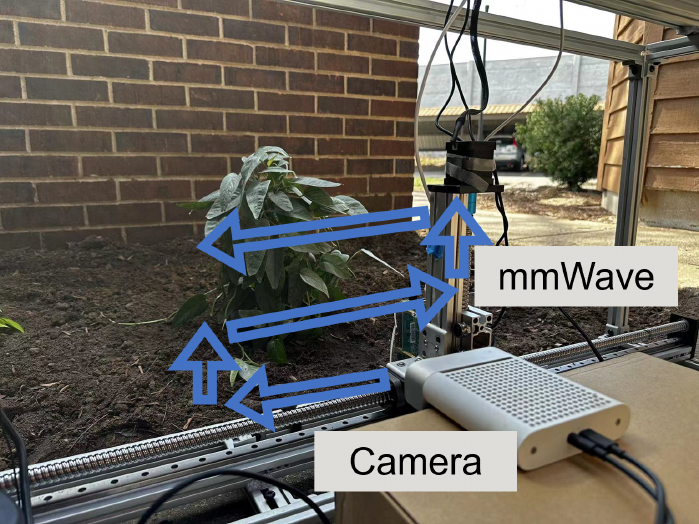}
   \caption{Lighting Condition}\label{fig:outdoor}
   \end{subfigure}
   \begin{subfigure}[b]{0.23\textwidth}
       \centering\includegraphics[width=\textwidth]{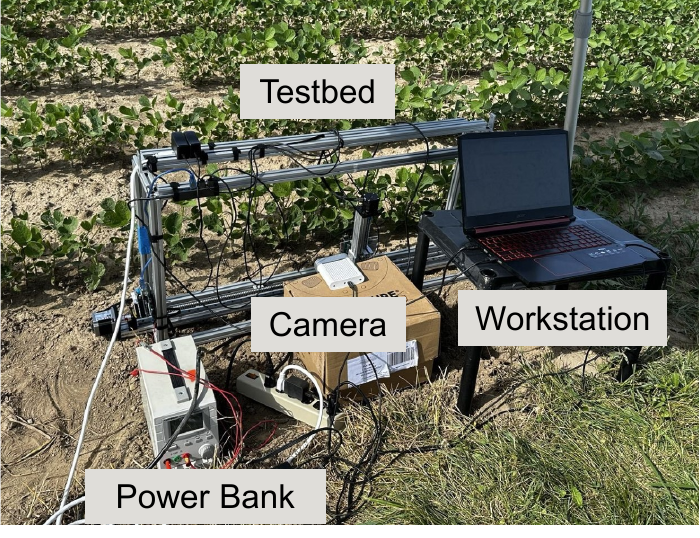}
   \caption{Plant Distance}\label{fig:farm}
   \end{subfigure}
   \label{fig:implementation}
    \caption{Data collection setup in indoor, outdoor, and farm. }
\end{figure}

As illustrated in Figure~\ref{fig:radar}, our SAR imaging system is equipped with a two-axis mechanical scanner optimized for high-speed data acquisition. 
The scanner is meticulously calibrated for plant-scale analysis, featuring a horizontal range of 150 mm and a vertical range of 100 mm. 
The mmWave radar mounted on this scanning platform is a Texas Instruments (TI) IWR1642 radar module \cite{mmwave-radar}, which operates in the $77$–$81 GHz$ frequency band. Each chirp signal consists of 256 sampling points, and its frequency will increase from $f_0 = 77GHz$
to $f_T = 80.99GHz$ with the bandwidth $B = 3.99GHz$ and frequency slope $k = 70.295 MHz/\mu$.
This radar captures raw mmWave reflections from plant surfaces, serving as the core sensing unit. 
Signal collection is managed by the TI DCA1000EVM~\cite{mmwave-receiver}, which interfaces with the radar to collect and store signals for subsequent imaging and analysis.

\begin{figure}[t!]

\begin{minipage}[b]{.6\linewidth}
  \centering
  \centerline{\includegraphics[width=4.7cm]{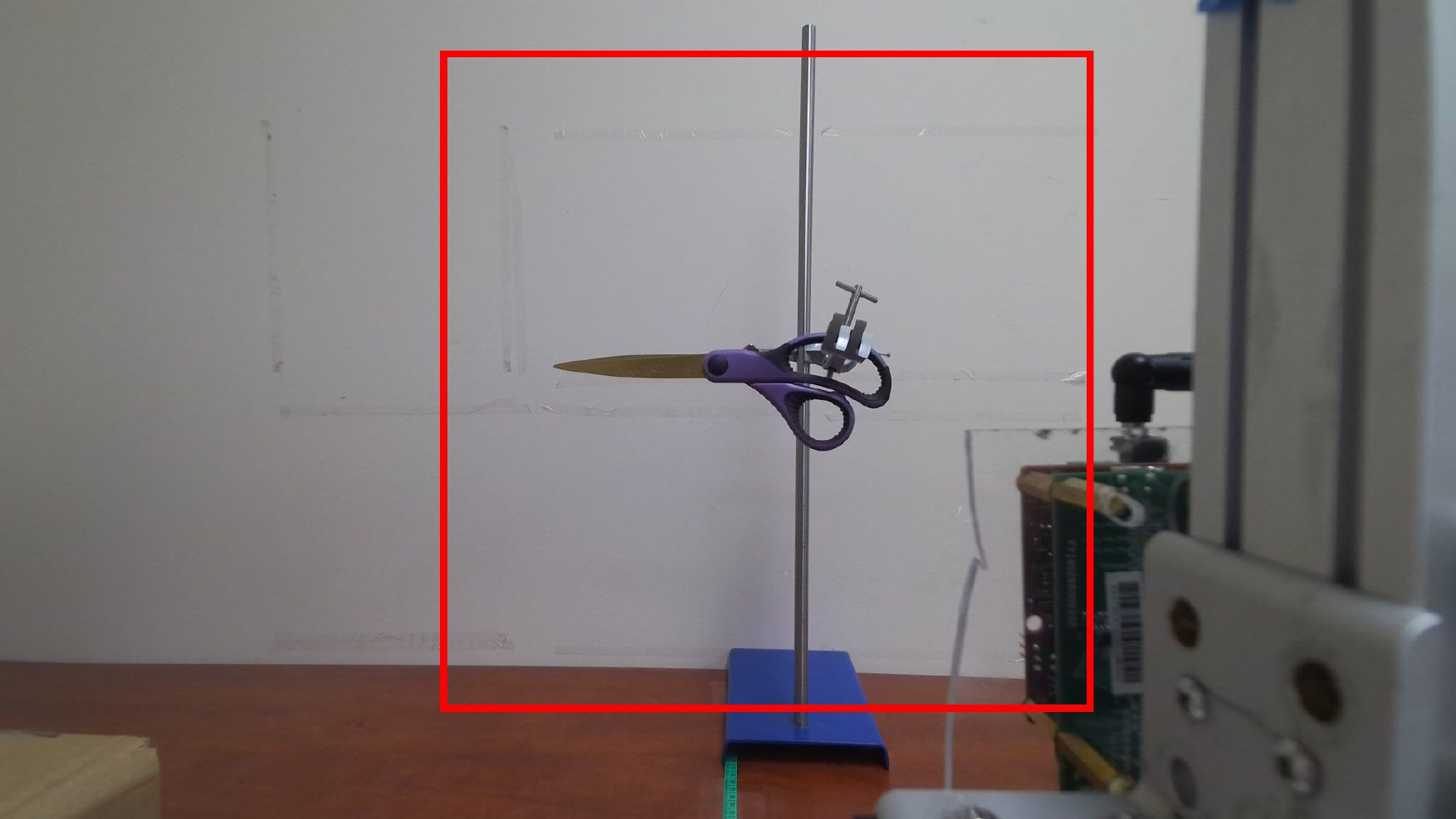}}
  \centerline{RGB camera Image}\medskip
\end{minipage}
\hfill
\begin{minipage}[b]{.39\linewidth}
  \centering
  \centerline{\includegraphics[width=2.7cm]{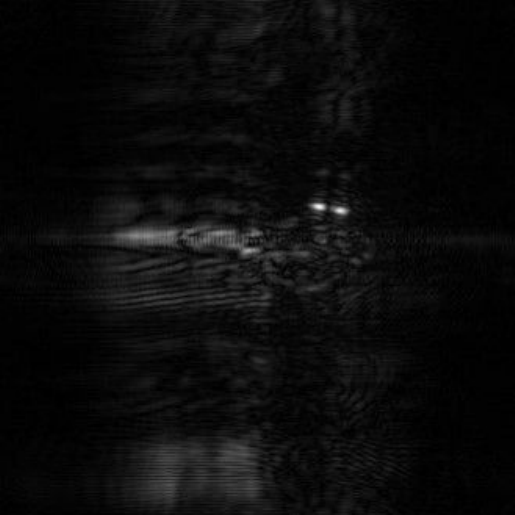}}
  \centerline{SAR Image}\medskip
\end{minipage}

\caption{Calibration of the RGB camera and SAR with the scissor. The red rectangle is the shared field of view.}
\label{fig:calibration}
\end{figure}

To complement the SAR system, our prototype also integrates a camera imaging module using an Azure Kinect sensor~\cite{kinect}, strategically positioned at the center of the scanning area.
SAR and camera systems are carefully calibrated to share a common field of view, as shown in Figure \ref{fig:calibration}.
The fusion of SAR-based mmWave imaging and RGB camera imagery offers a comprehensive, multimodal representation of the leaf surface, enhancing detection accuracy and robustness.

\subsection{SAR Imaging}
SAR is a radar imaging technique that achieves high spatial resolution by synthesizing a large aperture through relative motion between the radar and the target \cite {meta2007signal}. 
In our system, we integrate SAR with FMCW radar, which emits chirped signals whose instantaneous frequency increases linearly over time \cite{yanik2019near, yanik2020development}. 
Upon receiving the backscattered signals, the system performs a dechirping process to extract the beat frequency, isolating key features indicative of target range and reflectivity. 
These signals are expressed in the wave number domain, where spatial and geometric relations between the radar and scatter points are accurately modeled. 
Using a range migration algorithm, we convert multi-static measurements into a monostatic equivalent via phase compensation, simplifying depth-based imaging. 
The final 2D image reconstruction is achieved through an inverse Fourier transform of the backscattered data, supported by Weyl’s representation theorem\cite{weyl1919ausbreitung}, which approximates spherical waves as a superposition of plane waves. 
We implement and optimize the entire imaging pipeline in Python, leveraging GPU acceleration to support real-time and large-scale processing.

\subsection{Dataset}
Our indoor experiments spanned a six-month data collection period across five diverse plant species, selected to capture variations in leaf size, orientation, and structural complexity. 
Throughout this period, the plants exhibited distinct growth patterns, spatial arrangements, and leaf distributions, contributing to the dataset’s richness and diversity. 
The experimental setup positioned each plant at a distance of $200$–$500 mm$ from the mmWave radar, and data collection focused on two well-defined states: fully dry and fully saturated leaf surfaces. 
In total, we compiled a dataset comprising $292$ multimodal sample pairs. 
This includes $268$ pairs collected under controlled indoor conditions, as shown in Figure \ref{fig:indoor}, and $24$ pairs from outdoor environments, as shown in Figure \ref{fig:outdoor}, which exhibit dynamic environmental variability, such as fluctuating lighting and weather. This comprehensive dataset enables robust model training and evaluation under both controlled and real-world conditions.

\subsection{File Structure}
Inside the dataset, there exist four folders: source code, RGB image, SAR image, and their corresponding raw data.
The source code includes a SAR imaging algorithm written in Python, as well as a GPU-accelerated version.
Additionally, for all datasets, all file names within each folder are identical, representing samples captured simultaneously.
\subsection{Dataset Naming Convention}

The dataset adopts a structured naming convention to facilitate efficient sample identification and retrieval. Each filename consists of multiple segments separated by underscores (\_), encoding the following metadata:

\begin{itemize}
    \item \textbf{Group Status}: Indicates the leaf condition, where \texttt{0} denotes a dry sample and \texttt{1} denotes a wet sample.
    \item \textbf{Sampling Date}: Represents the collection date in \texttt{MMDD} format.
    \item \textbf{Sensor Distance}: Specifies the closest distance from the plant to the radar sensor, in millimeters.
    \item \textbf{Sample Index}: A unique identifier for the specific sample within the group.
    \item \textbf{Cross-Section Distance} (for SAR images): Indicates the cross-sectional depth from the radar sensor where the SAR image was captured.
\end{itemize}

For example, the file path:
\begin{center}
\texttt{0\_0119\_200\_1\_200.jpg}
\end{center}
corresponds to a dry leaf sample (group \texttt{0}) collected on January 19 (\texttt{0119}), with the plant positioned 200~mm from the radar. It is the first sample (\texttt{1}) in that group, and the SAR cross-section was recorded at a depth of 200~mm from the radar.

In addition to processed SAR and RGB images, each sample directory contains the corresponding raw data files, allowing for custom signal processing and further development of the imaging algorithm.

%% file: content/4_testing.tex
\subsection{Testing Algorithm}
After constructing our dataset, we evaluate its utility through a comprehensive benchmarking process centered on the Hydra framework~\cite{hydra24liu}, which employs a multimodal data fusion algorithm for leaf wetness detection.
 Our primary evaluation focuses on the accuracy of the classification.
 Specifically, the model's ability to determine whether a leaf surface is wet or dry. 
 The Hydra algorithm uses a two-stage fusion pipeline. 
 First, it performs depth-aware fusion by aligning SAR images captured at multiple depths with high-resolution RGB imagery and extracts features through a Convolution Neural Network (CNN). 
 In the second stage, Hydra utilizes a transformer-based encoder that models sequential relationships across different SAR depths using depth-aware positional encoding and multi-head attention. 
 This enables the system to construct a coherent 3D understanding of plant surfaces, allowing robust classification under varying conditions.


%% file: content/5_evaluation.tex
\begin{figure}[!t]
   \centering
   \begin{subfigure}[b]{0.23\textwidth}
       \centering\includegraphics[width=\textwidth]{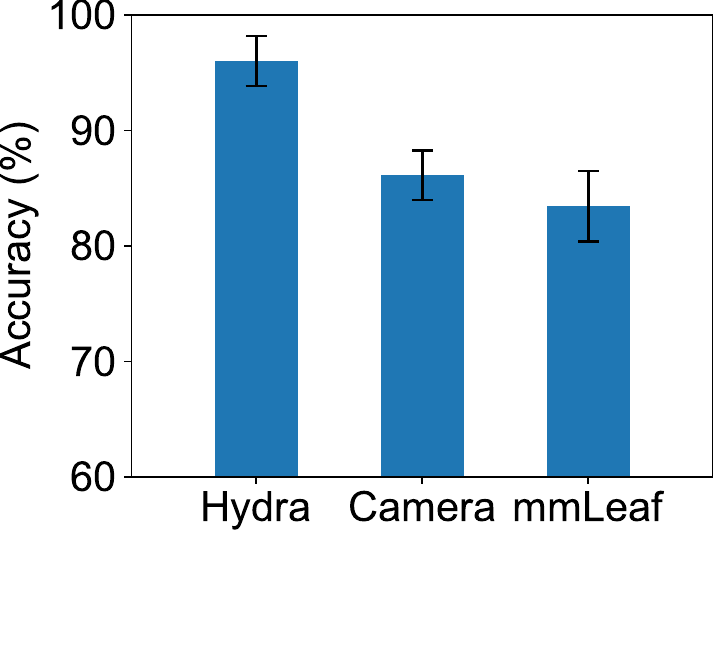}
   \caption{LWD Accuracy}\label{fig:overall-acc} 
   \captionsetup{skip=-1pt}
   \end{subfigure}
   \begin{subfigure}[b]{0.23\textwidth}
       \centering\includegraphics[width=\textwidth]{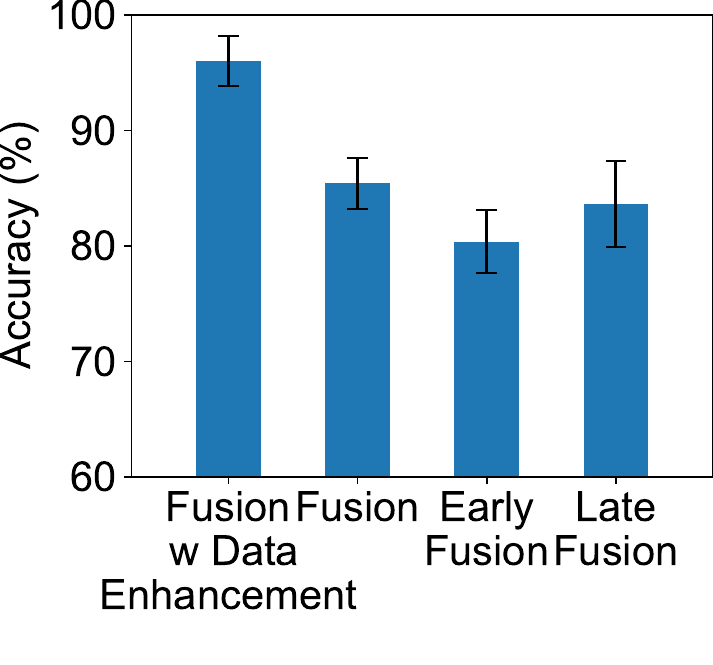}
   \caption{Fusion Performance}\label{fig:fusion-acc}
   \captionsetup{skip=-1pt}
   \end{subfigure}
   \begin{subfigure}[b]{0.23\textwidth}
       \centering\includegraphics[width=\textwidth]{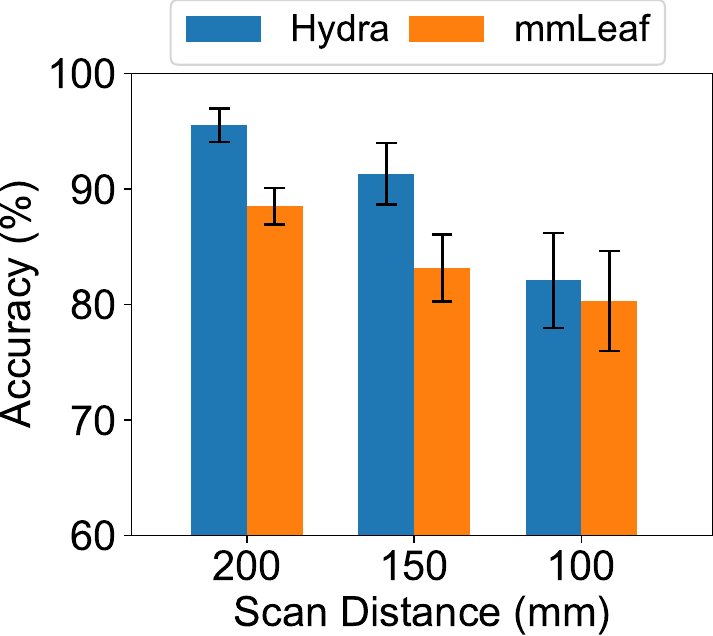}
   \caption{Scan Dist. Performance}
   \captionsetup{skip=-0.5pt}
   \label{fig:scan-distance}
   \end{subfigure}
   \begin{subfigure}[b]{0.23\textwidth}
       \centering\includegraphics[width=\textwidth]{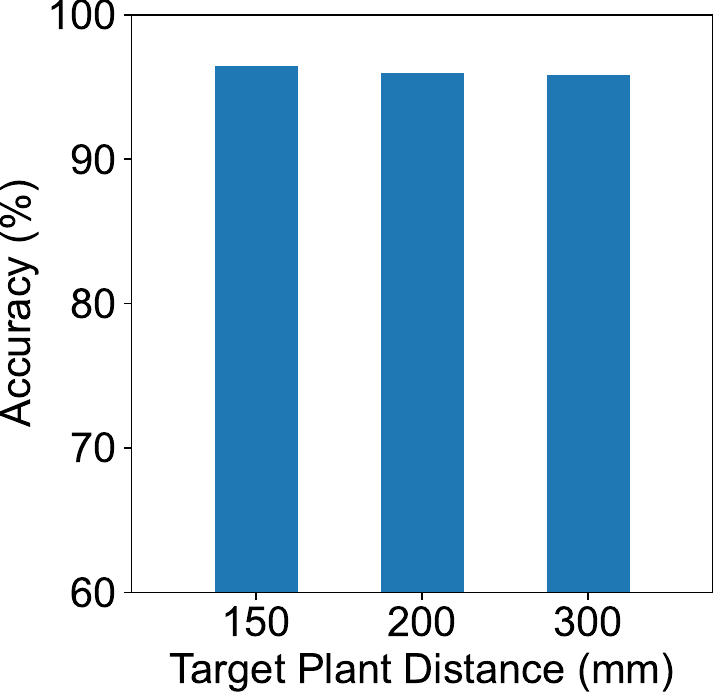}
   \caption{Target Distance Performance}\label{fig:distance}
   \captionsetup{skip=-0.5pt}
   \end{subfigure}
   \label{fig:overall-eval}
    \caption{Overall performance for Hydra LWD.}
\end{figure}

\section{Evaluation}

We conduct a comprehensive evaluation of Hydra to assess its accuracy, robustness, and efficiency in detecting leaf wetness under diverse environmental conditions. 
Our evaluation focuses on distinguishing between wet and dry leaves. 
As shown in Figure \ref{fig:overall-acc}, Hydra achieves an accuracy of $96\% \pm 2.14\%$ in indoor scenarios and maintains robust performance with approximately $90\%$ accuracy in real-world farm environments. 
For single modality baselines, the camera-only model achieves $86.13\% \pm 2.13\%$ accuracy, while the SAR-only baseline reaches $83.43\% \pm 3.05\%$. 

We also compare several multimodal fusion strategies shown in Figure \ref{fig:fusion-acc}. 
Hydra's proposed depth-aware fusion achieves $96\% \pm 2.14\%$, which improves to $85.41\% \pm 2.24\%$ with our data enhancement techniques. 
In contrast, early fusion and late fusion methods yield $80.38\% \pm 1.84\%$ and $83.63\% \pm 1.84\%$ accuracy, respectively. 

Our evaluation demonstrates that SAR imaging maintains high accuracy across a range of scan distances shown in Figure \ref{fig:scan-distance}. 
Shorter scan distances result in reduced resolution and narrower coverage, leading to performance degradation.
For the scan distance longer or equal to 150 mm, with the advantage of the multi-modality, the accuracy is above 90\%, which is $95.52\% \pm 1.46\%, 91.32\% \pm 2.65\% $ and $82.1\% \pm 4.13\%$ for 200mm, 150mm and 100mm, respectively.
They improved from single modality accuracy which is $88.32\% \pm 1.58\%, 83.17\% \pm 2.9\% $ and $80.3\% \pm 4.32\%$. 
Hydra consistently outperforms mmLeaf by preserving detection accuracy and reducing scanning inefficiency by 25\%. 
These results highlight the effectiveness of multi-modality in mitigating the trade-offs between scan efficiency and detection precision.

To evaluate the impact of imaging configuration, we analyze Hydra's performance across varying SAR scan distances as shown in Figure \ref{fig:scan-distance}.
Hydra maintains high accuracy at wider scan distances: $95.43\% \pm 1.47\%$ at 200~mm, $94.68\% \pm 1.97\%$ at 175~mm, and $93.38\% \pm 2.56\%$ at 150~mm. 
However, performance degrades at shorter distances, dropping to $89.28\% \pm 4.05\%$ at 125 mm and $84.10\% \pm 4.53\%$ at 100 mm due to reduced resolution and a limited field of view.
These results underscore Hydra's superior capability in leveraging multi-modal features and optimizing SAR imaging for accurate and efficient leaf wetness detection.

%% file: content/6_conclusion.tex
\section{Related Work}
\noindent \textbf{Artificial Intelligence in Agricultural IoT.}  
The convergence of artificial intelligence and the Internet of Things (AIoT) has transformed agricultural practices, enabling intelligent, scalable, and connected farm management~\cite{aiot2024liu}. Long-range communication technologies, such as LoRa and satellite-based networking, improve data coverage and reliability for rural applications~\cite{dong2024demeter, dong2024sate, Adwait2018lora, Akshay2024lora, AeroEcho25Ren}. On the sensing parts, modalities like RF and VNIR imaging facilitate soil health monitoring~\cite{soil2024wang}. At the same time, multi-modality, machine learning, and reinforcement learning approaches have been employed to optimize object detection in foliage, disease management, irrigation, and resource allocation~\cite{cardamis2024leafeon, ding2022irrigation, wang2022detection, gangeofl}, contributing to the development of sustainable and data-driven agricultural ecosystems.

\section{Conclusion}
This dataset paper presents a comprehensive and multi-modal dataset that addresses critical gaps in current leaf wetness detection research. By combining mmWave SAR and RGB imaging across a diverse set of plant species and environmental conditions, we enable rigorous benchmarking of machine learning models under both controlled and real-world settings. Our evaluations using the Hydra architecture demonstrate the value of depth-aware multi-modal fusion and provide clear baselines for camera-only, SAR-only, and hybrid models. We further analyze the impact of fusion strategy and SAR scan distance, highlighting key trade-offs in detection accuracy and imaging efficiency. This dataset not only advances the state of the art in precision agriculture sensing but also offers a foundational resource for the machine learning community to explore robust, explainable, and efficient multi-modal systems.

\vspace{1mm}
\section{Acknowledgement}
\label{sec:ack}
\noindent
This work was partially supported by NSF CAREER Award 2338976.

%% file: main.bbl
\begin{thebibliography}{10}

\bibitem{hydra24liu}
Yimeng Liu, Maolin Gan, Huaili Zeng, Li~Liu, Younsuk Dong, and Zhichao Cao,
\newblock ``Hydra: Accurate multi-modal leaf wetness sensing with mm-wave and camera fusion,''
\newblock in {\em Proceedings of ACM MobiCom}, 2024.

\bibitem{worldbankAgri}
{THE WORLD BANK},
\newblock ``Agriculture and food,'' \url{https://www.worldbank.org/en/topic/agriculture/overview}, 2024,
\newblock Accessed: 2024-03-14.

\bibitem{Singh2023Climate}
Brajesh~K Singh, Manuel Delgado-Baquerizo, Eleonora Egidi, Emilio Guirado, Jan~E Leach, Hongwei Liu, and Pankaj Trivedi,
\newblock ``Climate change impacts on plant pathogens, food security and paths forward,''
\newblock {\em Nature Reviews Microbiology}, vol. 21, no. 10, pp. 640--656, 2023.

\bibitem{plants12152800}
Concepció Moragrega and Isidre Llorente,
\newblock ``Effects of leaf wetness duration, temperature, and host phenological stage on infection of walnut by xanthomonas arboricola pv. juglandis,''
\newblock {\em MDPI Plants}, vol. 12, no. 15, 2023.

\bibitem{huber1992modeling}
L~Huber and TJ~Gillespie,
\newblock ``Modeling leaf wetness in relation to plant disease epidemiology,''
\newblock {\em Annual review of phytopathology}, vol. 30, no. 1, pp. 553--577, 1992.

\bibitem{rowlandson2015reconsidering}
Tracy Rowlandson, Mark Gleason, Paulo Sentelhas, Terry Gillespie, Carla Thomas, and Brian Hornbuckle,
\newblock ``Reconsidering leaf wetness duration determination for plant disease management,''
\newblock {\em Plant Disease}, vol. 99, no. 3, pp. 310--319, 2015.

\bibitem{mackenzie2012use}
SJ~MacKenzie and NA~Peres,
\newblock ``Use of leaf wetness and temperature to time fungicide applications to control anthracnose fruit rot of strawberry in florida,''
\newblock {\em Plant disease}, vol. 96, no. 4, pp. 522--528, 2012.

\bibitem{gubler1999control}
WD~Gubler, MR~Rademacher, SJ~Vasquez, and CS~Thomas,
\newblock ``Control of powdery mildew using the uc davis powdery mildew risk index,''
\newblock {\em APSnet Feature. Published online. The American Phytopathological Society, St. Paul, MN}, 1999.

\bibitem{wu2001validation}
BM~Wu, KV~Subbarao, AHC~van Bruggen, and GGH Pennings,
\newblock ``Validation of weather and leaf wetness forecasts for a lettuce downy mildew warning system,''
\newblock {\em Canadian journal of plant pathology}, vol. 23, no. 4, pp. 371--383, 2001.

\bibitem{duvdevani1947optical}
S~Duvdevani,
\newblock ``An optical method of dew estimation,''
\newblock {\em Quarterly Journal of the Royal Meteorological Society}, vol. 73, no. 317-318, pp. 282--296, 1947.

\bibitem{gan2023poster}
Maolin Gan, Yimeng Liu, Li~Liu, Chenshu Wu, Younsuk Dong, Huacheng Zeng, and Zhichao Cao,
\newblock ``Poster: mmleaf: Versatile leaf wetness detection via mmwave sensing,''
\newblock in {\em Proceedings of ACM MobiSys}, 2023.

\bibitem{PHYTOS31}
{METER Group},
\newblock ``{PHYTOS 31 Manual Web},'' \url{http://library.metergroup.com/Manuals/20434_PHYTOS31_Manual_Web.pdf}, 2021,
\newblock Accessed: Nov 21, 2022.

\bibitem{nguyen2023bio}
Brian~H Nguyen, Gregory~S Gilbert, and Marco Rolandi,
\newblock ``A bio-mimetic leaf wetness sensor from replica molding of leaves,''
\newblock {\em Advanced Sensor Research}, vol. 2, no. 6, pp. 2200033, 2023.

\bibitem{Adonis25Liu}
Yimeng Liu, Maolin Gan, Gen Li, Younsuk Dong, and Zhichao Cao,
\newblock ``Adonis: Neural-enhanced fine-grained leaf wetness sensing with efficient mmwave imaging,''
\newblock in {\em Proceedings of IEEE INFOCOM}, 2025.

\bibitem{Proteus25Liu}
Yimeng Liu, Maolin Gan, Huaili Zeng, Yidong Ren, Gen Li, Younsuk Dong, Xiaobo Tan, and Zhichao Cao,
\newblock ``Proteus: : Enhanced mmwave leaf wetness detection with cross-modality knowledge transfer,''
\newblock in {\em Proceedings of ACM SenSys}, 2025.

\bibitem{mmwave-radar}
{Texas Instruments},
\newblock ``Iwr1642,'' \url{https://www.ti.com/product/IWR1642}, 2024,
\newblock Accessed: 2024-Oct-28.

\bibitem{mmwave-receiver}
Texas Instruments,
\newblock ``Dca1000evm,'' \url{https://www.ti.com/tool/DCA1000EVM}, 2024,
\newblock Accessed: 2024-Oct-28.

\bibitem{kinect}
{Microsoft},
\newblock ``Azure kinect dk,'' \url{https://azure.microsoft.com/en-us/products/kinect-dk}, 2024,
\newblock Accessed: 2024-10-28.

\bibitem{meta2007signal}
Adriano Meta, Peter Hoogeboom, and Leo~P Ligthart,
\newblock ``Signal processing for fmcw sar,''
\newblock {\em IEEE Transactions on Geoscience and Remote Sensing}, vol. 45, no. 11, pp. 3519--3532, 2007.

\bibitem{yanik2019near}
Muhammet~Emin Yanik and Murat Torlak,
\newblock ``Near-field mimo-sar millimeter-wave imaging with sparsely sampled aperture data,''
\newblock {\em IEEE Access}, vol. 7, pp. 31801--31819, 2019.

\bibitem{yanik2020development}
Muhammet~Emin Yanik, Dan Wang, and Murat Torlak,
\newblock ``Development and demonstration of mimo-sar mmwave imaging testbeds,''
\newblock {\em IEEE Access}, vol. 8, pp. 126019--126038, 2020.

\bibitem{weyl1919ausbreitung}
H.~Weyl,
\newblock ``Ausbreitung elektromagnetischer wellen über einem ebenen leiter,''
\newblock {\em Annalen der Physik}, vol. 365, no. 21, pp. 481--500, 1919.

\bibitem{aiot2024liu}
Shakhrul~Iman Siam, Hyunho Ahn, Li~Liu, Samiul Alam, Hui Shen, Zhichao Cao, Ness Shroff, Bhaskar Krishnamachari, Mani Srivastava, and Mi~Zhang,
\newblock ``Artificial intelligence of things: A survey,''
\newblock {\em ACM Trans. Sen. Netw.}, Aug. 2024,
\newblock Just Accepted.

\bibitem{dong2024demeter}
Yidong Ren, Wei Sun, Jialuo Du, Huaili Zeng, Younsuk Dong, Mi~Zhang, Shigang Chen, Yunhao Liu, Tianxing Li, and Zhichao Cao,
\newblock ``Demeter: Reliable cross-soil lpwan with low-cost signal polarization alignment,''
\newblock in {\em Proceedings of ACM MobiCom}, 2024.

\bibitem{dong2024sate}
Yidong Ren, Amalinda Gamage, Li~Liu, Mo~Li, Shigang Chen, Younsuk Dong, and Zhichao Cao,
\newblock ``Sateriot: High-performance ground-space networking for rural iot,''
\newblock in {\em Proceedings of ACM MobiCom}, 2024.

\bibitem{Adwait2018lora}
Adwait Dongare, Revathy Narayanan, Akshay Gadre, Anh Luong, Artur Balanuta, Swarun Kumar, Bob Iannucci, and Anthony Rowe,
\newblock ``Charm: Exploiting geographical diversity through coherent combining in low-power wide-area networks,''
\newblock in {\em 2018 17th ACM/IEEE International Conference on Information Processing in Sensor Networks (IPSN)}, 2018, pp. 60--71.

\bibitem{Akshay2024lora}
Akshay Gadre, Zachary Machester, and Swarun Kumar,
\newblock ``Adapting lora ground stations for low-latency imaging and inference from lora-enabled cubesats,''
\newblock {\em ACM Trans. Sen. Netw.}, vol. 20, no. 5, July 2024.

\bibitem{AeroEcho25Ren}
Yidong Ren, Gen Li, Yimeng Liu, Younsuk Dong, and Zhichao Cao,
\newblock ``Aeroecho: Towards agricultural low-power wide-area backscatter with aerial excitation source,''
\newblock in {\em Proceedings of IEEE INFOCOM}, 2025.

\bibitem{soil2024wang}
Juexing Wang, Yuda Feng, Gouree Kumbhar, Guangjing Wang, Qiben Yan, Qingxu Jin, Robert~C. Ferrier, Jie Xiong, and Tianxing Li,
\newblock ``Soilcares: Towards low-cost soil macronutrients and moisture monitoring using rf-vnir sensing,''
\newblock in {\em Proceedings of ACM MobiSys}, 2024.

\bibitem{cardamis2024leafeon}
Mark Cardamis, Hong Jia, Hao Qian, Wenyao Chen, Yihe Yan, Oula Ghannoum, Aaron Quigley, Chung~Tung Chou, and Wen Hu,
\newblock ``Leafeon: Towards accurate, robust and low-cost leaf water content sensing using mmwave radar,''
\newblock {\em arXiv preprint arXiv:2410.03680}, 2024.

\bibitem{ding2022irrigation}
Xianzhong Ding and Wan Du,
\newblock ``Drlic: Deep reinforcement learning for irrigation control,''
\newblock in {\em 2022 21st ACM/IEEE International Conference on Information Processing in Sensor Networks (IPSN)}, 2022, pp. 41--53.

\bibitem{wang2022detection}
Ruihao Wang, Yimeng Liu, and Rolf M{\"u}ller,
\newblock ``Detection of passageways in natural foliage using biomimetic sonar,''
\newblock {\em Bioinspiration \& Biomimetics}, vol. 17, no. 5, pp. 056009, 2022.

\bibitem{gangeofl}
Maolin Gan, Lanpeng Li, Samiul Alam, Li~Liu, Luyang Liu, Mi~Zhang, and Zhichao Cao,
\newblock ``Geofl: A framework for efficient geo-distributed cross-device federated learning,''
\newblock in {\em Proceedings of IEEE INFOCOM}, 2025.

\end{thebibliography}
